\documentclass[letterpaper, 10 pt, conference]{ieeeconf}  
\usepackage{amsmath,amsfonts}

\usepackage{graphics}
\usepackage{stfloats}
\usepackage{epsfig}
\usepackage{times}
\usepackage{amssymb}
\usepackage{amsmath}
\usepackage{indentfirst}
\usepackage{epstopdf}
\usepackage{multirow}
\usepackage{pdfsync}
\usepackage{float}
\usepackage{color}
\usepackage{enumerate}
\usepackage{subfigure}
\usepackage{flushend}
\IEEEoverridecommandlockouts                            
\usepackage{graphicx}      

\usepackage{amsfonts}
\DeclareMathSymbol{\shortminus}{\mathbin}{AMSa}{"39}
\let\labelindent\relax
\usepackage{enumitem}
\usepackage{multirow}
\usepackage{tikz}
\usepackage{booktabs}
\usepackage{array}
\usepackage{float}
\usepackage{amsmath}
\usepackage{epsfig}
\usepackage{arydshln}
\usepackage{algpseudocode}
\usepackage{verbatim}
\usepackage{subfigure}
\usepackage{enumerate}
\usepackage{rotating}
\usepackage{color}
\usepackage{flushend}
\usepackage{mathrsfs}
\usepackage{placeins}
\usepackage[ruled,vlined,linesnumbered]{algorithm2e}
\usepackage{float}
\usepackage{soul}
\usepackage{fancyhdr}

\usepackage{url}
\usepackage{makecell} 
\usepackage{mathtools}
\usepackage{amsmath} 
\usepackage[para,online,flushleft]{threeparttable}
\definecolor{myyellow}{RGB}{205,192,176} 
\definecolor{newcolor}{rgb}{0.21,0.49,0.66}
\definecolor{Rcolor}{rgb}{1,0,0}

\title{\bf Diff-CXR: Report-to-CXR generation through a disease-knowledge enhanced diffusion model} 

\author{Peng Huang, Bowen Guo, Shuyu Liang, Junhu Fu, Yuanyuan Wang and Yi Guo$^{\ast}$ \thanks{*Corresponding author {\tt\small  E-mail: guoyi@fudan.edu.cn}}
\thanks{This work was supported by the National Natural Science Foundation of China (Grant 62371139 and 82227803), and the Science and Technology Commission of Shanghai Municipality, China (Grant 22ZR1404800).}
\thanks{P. Huang
	{\tt\small  E-mail: 22210720084@m.fudan.edu.cn}.
}
}

\floatname{algorithm}{Procedure} 
\begin{document}
\maketitle
\begin{abstract}     
 Text-To-Image (TTI) generation is significant for controlled and diverse image generation with broad potential applications. Although current medical TTI methods have made some progress in report-to-Chest-Xray (CXR) generation, their generation performance may be limited due to the intrinsic characteristics of medical data. In this paper, we propose a novel disease-knowledge enhanced Diffusion-based TTI learning framework, named Diff-CXR, for medical report-to-CXR generation. First, to minimize the negative impacts of noisy data on generation, we devise a Latent Noise Filtering Strategy that gradually learns the general patterns of anomalies and removes them in the latent space. Then, an Adaptive Vision-Aware Textual Learning Strategy is designed to learn concise and important report embeddings in a domain-specific Vision-Language Model, providing textual guidance for Chest-Xray generation. Finally, by incorporating the general disease knowledge into the pretrained TTI model via a delicate control adapter, a disease-knowledge enhanced diffusion model is introduced to achieve realistic and precise report-to-CXR generation. Experimentally, our Diff-CXR outperforms previous SOTA medical TTI methods by 33.4\% / 8.0\% and 23.8\% / 56.4\% in the FID and mAUC score on MIMIC-CXR and IU-Xray, with the lowest computational complexity at 29.641 GFLOPs. Downstream experiments on three thorax disease classification benchmarks and one CXR-report generation benchmark demonstrate that Diff-CXR is effective in improving classical CXR analysis methods. Notably, models trained on the combination of 1\% real data and synthetic data can achieve a competitive mAUC score compared to models trained on all data, presenting promising clinical applications.
\end{abstract}


\section{Introduction} 
\subsection{Background and Motivation}
Image generation involves utilizing computer algorithms to produce novel images, aiming to generate realistic or stylized visual content based on specified conditions. It has gained popularity across diverse domains, encompassing super-resolution, style transfer, image reconstruction and restoration, text-to-image generation (TTI) et al \cite{MaPathSRGAN2020,MingliCTsynthesis2023,yeung2024sensorless,PatrickTransformerSynthesis, WU2024103284}. These achievements underscore its promising potential applications in fields such as art, medicine, and education.

In medical scene, dominant image generation methods focus on image-to-image generation, among which the most representative methods are Generative Adversarial Network (GAN) and its variants. As a generator endeavors to learn the distribution of real examples, GAN produces high-quality images to deceive the discriminator \cite{goodfellowGenerativeAdversarialNets2014}. Various architectures and losses based on GAN have been designed to generate high-quality medical images of different modalities and anatomical regions \cite{arjovskyWassersteinGenerativeAdversarial2017,MehdiCGAN2014}. However, the diversity and controllability of generated samples are limited due to the challenge of specifying nuanced visual attributes solely within image-to-image generation process.

Text-to-image generation emerges and provides a more intuitive and detailed way to specify visual attributes through descriptive language. It exhibits refined semantic control, enabling the generation of more diverse samples to supplement downstream tasks. Additionally, it contributes to the efficient data transmission and storage, as well as a swift image-retrieval system.

\begin{figure}[!t]
\centerline{\includegraphics[width=\columnwidth]{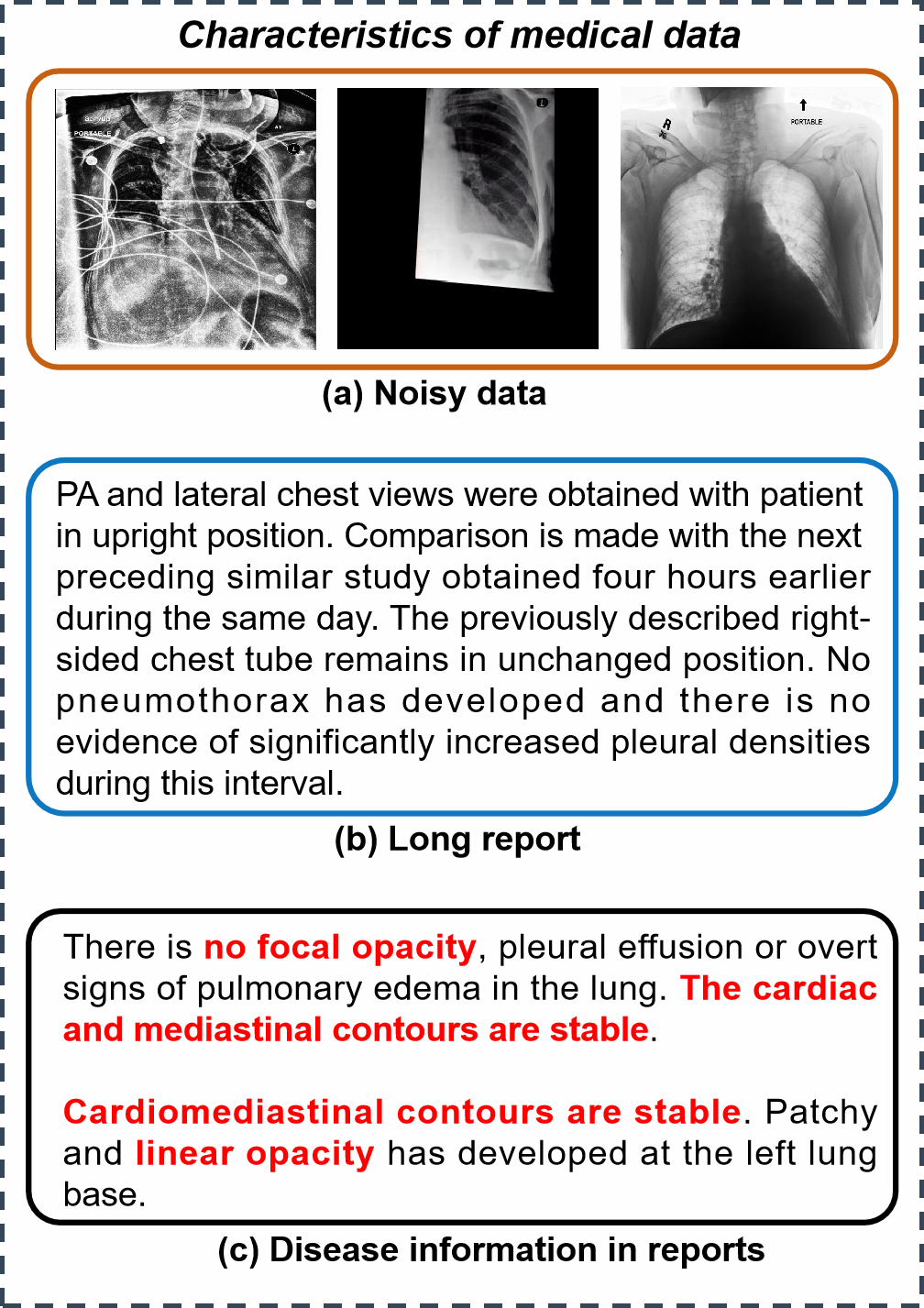}}
\caption{Illustration of characteristics of medical data which are essential to report-to-CXR generation process. In detail, (a) visualizes several noisy examples. (b) shows a long report example, which should be padded to meet the maximum textual token limits, 256, during the encoding process of VLMs. (c) illustrates that patients with different diseases may have same image manifestations. Disease information and imaging characteristics are highlighted in red for given reports.}
\label{fig1}
\end{figure}

Chest X-ray (CXR) is one of the most favored imaging modalities due to their rapid acquisition, non-invasive nature, and low-radiation exposure characteristics. Presently, several studies have made progress in medical TTI tasks, which generate CXRs from clinical reports. First, transformer-based autoregressive model has been successfully developed to generate images from medical reports with tailored architectures or training strategies \cite{lee2024vision, llmcxr2024}. Their practical application and further research are impeded by excessive parameters and high inference costs. Diffusion models can strike a balance between image generation performance, and model parameters. Current research mainly focuses on adapting pre-trained Stable Diffusion (SD) \cite{RobinStableDiffusion} to the medical domain through fine-tuning strategies \cite{PierreFewshotSD2022,RoentGen2022}. However, the generation perf r.ormance and efficiency of both autoregressive-based and diffusion-based TTI model are still limited due to the intrinsic characteristics of medical images and reports, as illustrated in \textcolor{newcolor}{Fig. \ref{fig1}}. Specifically, medical imaging datasets usually contain noisy samples \cite{Antanas2023media, ZhouAnomaly2021}. TTI model is prone to overfitting to these noisy patterns, thereby markedly compromising the authenticity and precision of the generated images. Next, to manage lengthy medical reports, domain-specific Vision-Language Model (VLM) often increases the maximum textual token length \cite{BiomedClip2023}. Textual embeddings with increased dimensions will lead to substantial climb in the computational complexities of the diffusion model. Third, medical reports typically detail diverse patient disease conditions and manifestations. Consequently, it is imperative to focus on the strategies for maintaining or emphasizing the disease representations during the diffusion process. In general, data curating, textual learning, and disease knowledge enhancement are essential to the report-to-CXR generation process.

This work builds upon \cite{huang2024chest}, which performance is still limited due to the characteristics of medical data. The key difference of this work is the proposal of a disease-knowledge enhanced diffusion-based medical TTI learning framework, named Diff-CXR, which fully explores the potential of generative foundational model in report-to-CXR generation. 

Our major contributions are summarized as follows:
\begin{itemize}
\item We present a novel disease-knowledge enhanced Diffusion-based report-to-CXR generation framework, Diff-CXR, which is capable of generating large amounts of CXRs encompassing 11 common lung diseases.

\item Diff-CXR employs a robust Latent Noise Filtering Strategy (LNFS) to efficiently eliminate the noisy data, especially the blurred ones near the decision boundary, within the latent space of a powerful pretrained autoencoder, following a coarse-to-fine manner.

\item The adaptive vision-aware textual learning strategy (AVA-TLS) is designed to endow the large domain-specific VLM to explicitly model the context relationship between different tokens and dynamically learn concise and important report embeddings. Then, we devise a disease knowledge injection mechanism (DKIM) to incorporate disease knowledge into the pretrained TTI model via a delicate control adapter, to strengthen the disease representation in the textual embeddings and refine the diffusion process.

\item The effectiveness is extensively confirmed through a sequence of ablation and comparison studies. The results indicate that our Diff-CXR significantly outperforms previous SOTA medical TTI methods on two widely used benchmark datasets. Downstream experiments on three thorax disease classification tasks and one CXR-report generation task further prove the effectiveness and versatility of our method.
\end{itemize}

\begin{figure*}
\centerline{\includegraphics[width=1.8\columnwidth]{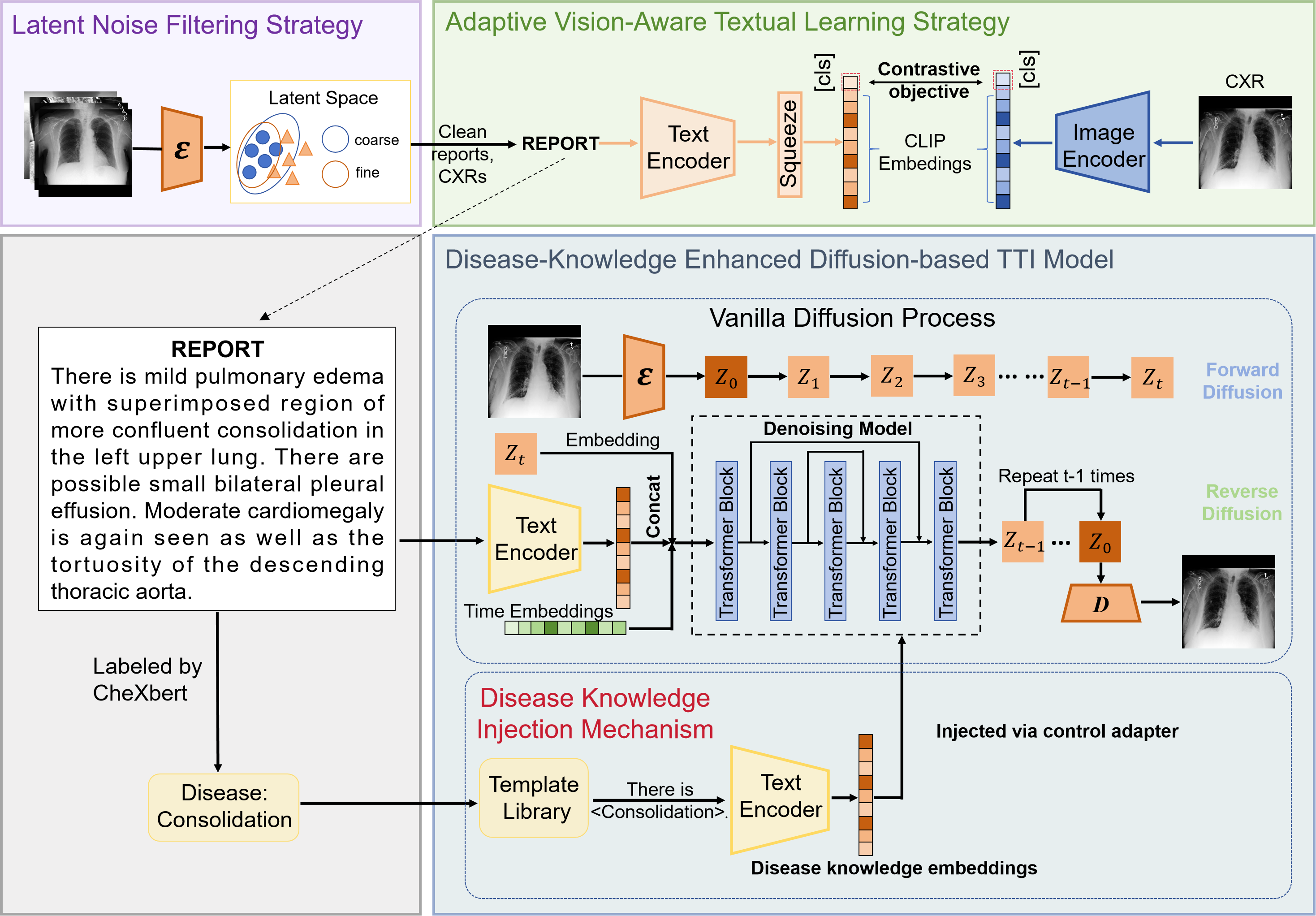}}
\caption{The overview of Diff-CXR. Given the training set, the latent noise filtering strategy effectively removes those noisy images with their reports in the latent space in a coarse-to-fine manner. Then, within the pruned dataset, the adaptive vision-aware textual learning strategy prompts the domain-specific language model to learn visually relevant and concise textual guidance for image generation. Finally, the disease-knowledge enhanced diffusion model is trained in two stages. In the vanilla diffusion process, only textual embeddings will be conditioned on the diffusion model. The disease-specific knowledge is further extracted and injected via a control adapter to refine generation results gradually.}
\label{fig2}   
\end{figure*}

\subsection{Related Works}
\subsubsection{Text-to-image generation}
Amidst the popularity of diffusion models, text-to-image generation has attained remarkable strides in natural domain \cite{RobinStableDiffusion,RameshDiffusion2022, UVIT2023, zhang2023adding}. The most representative diffusion-based TTI method is Stable Diffusion \cite{RobinStableDiffusion}, which scales up the latent diffusion model by increasing both the model size and data scale. It successfully achieves the trade-off between the computational complexity and the generation capability of diffusion models. Alternatively, U-ViT, a simpler and more general vision transformer architecture, is proposed, which treats time, image and text embeddings as tokens 
 \cite{UVIT2023}. U-ViT achieves better generation performance in TTI task and has become the mainstream architecture.

Several attempts have been made to perform TTI in medical domain in recent years, which can be divided into two groups, including transformer-based Autoregressive model and Diffusion-based model. Both UniXGen \cite{lee2024vision} and LLM-CXR \cite{llmcxr2024} follow the typical Autoregressive TTI pipeline, which predicts image tokens based on report embeddings and maps the image tokens to the image-level outputs with a VQ-GAN \cite{PatrickTransformerSynthesis} pre-trained in the medical domain. Although they can generate relatively realistic images from medical reports, extensive model parameters and high inference costs pose challenges for practical applications. Recently, two studies have carefully investigated the finetuning strategies to adapt Stable Diffusion to medical domain. \cite{PierreFewshotSD2022} shows the feasibility of finetuning Stable Diffusion to generate medical images in a few-short manner. RoentGen further explores the finetuning strategies and demonstrates that finetuning both UNet denoising model and CLIP-based text encoder of Stable Diffusion obtains the best generation results \cite{RoentGen2022}. RoentGen achieves relatively lower computational complexity, but the generation quality still lags behind autoregressive models. In this paper, we find that noisy images heavily impede the generation performance of diffusion-based TTI model and disease descriptions in medical reports are essential to the quality of medical TTI generation.

\subsubsection{Domain-specific VLM in medical domain}
To provide textual guidance for image generation, domain-specific Vision Language Model (VLM) is introduced to project clinical reports into meaningful conditional embeddings, which have shown impressive understanding abilities for both medical images and reports. PubMedCLIP directly finetunes the original CLIP \cite{radford2021learning} based on PubMed articles, achieving the state-of-the-art performance in Medical Visual Question Answering \cite{Pubmedclip2023}. Given that reports and images from different patients may share the same semantics, MedCLIP incorporates medical knowledge into the contrastive learning process to eliminate those false negatives \cite{MedClip2022}. It successfully outperforms previous best methods on zero-shot prediction, supervised classification, and image-text retrieval. However, both PubMedCLIP and MedCLIP are restricted to limited data and original CLIP-based architecture, further degrading their generalization abilities and applications in the medical domain. In this case, BioMedCLIP creates a novel dataset named PMC-15M, which is two orders of magnitude larger than existing biomedical multimodal datasets, to learn the unique characteristics of medical images and texts \cite{BiomedClip2023}. In addition, it employs a domain-specific language model, PubMedBERT \cite{GuyuCLIP2021} for the text encoder, and adjusts the tokenizer and context size to model the typically longer biomedical captions. However, this results in the increasing feature dimension in textual embeddings and the high computation complexity of diffusion model.

\begin{figure*}
\centerline{\includegraphics[width=1.8\columnwidth]{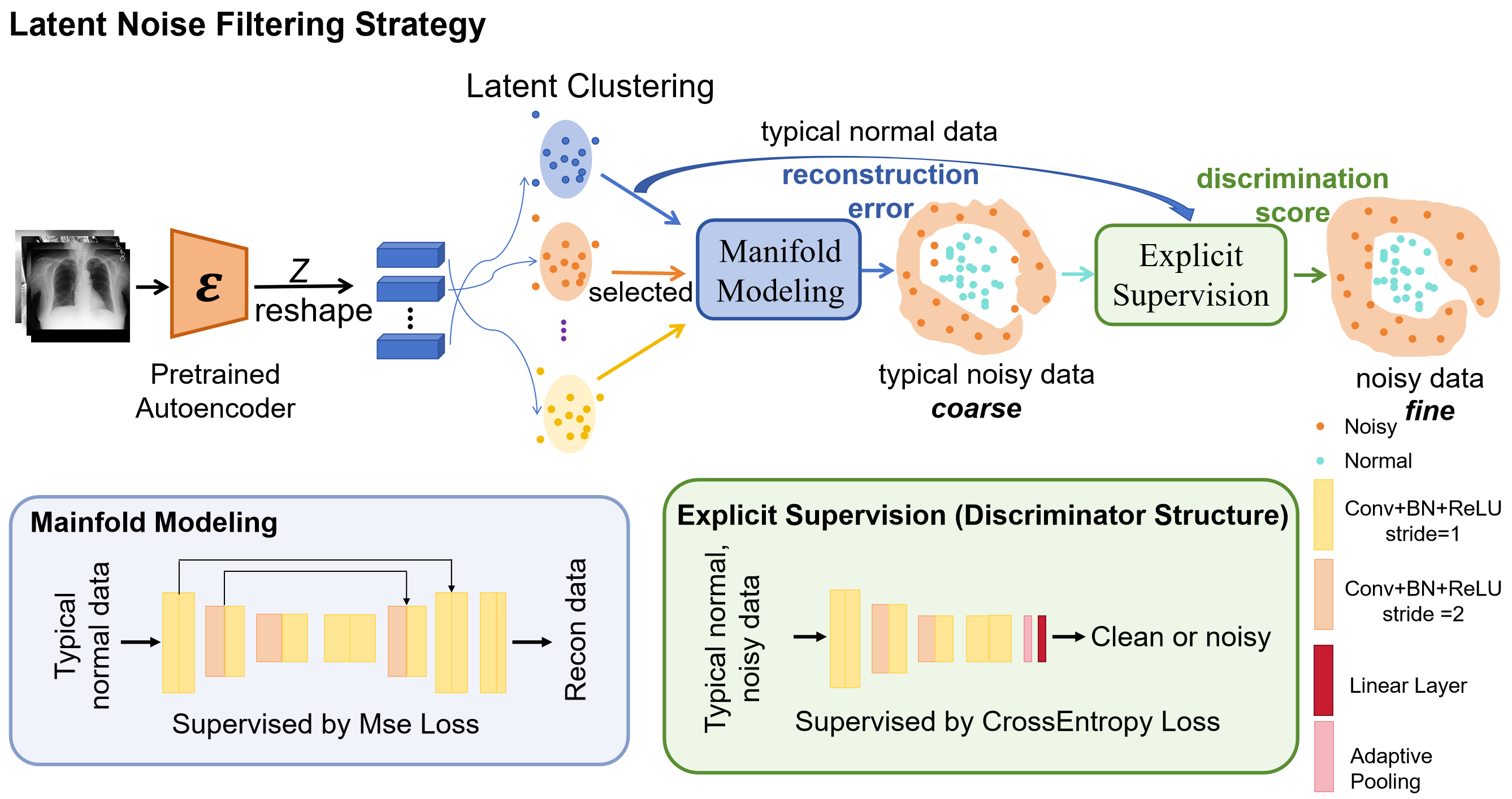}}
\caption{Diagram of the latent noise filtering strategy. Latent clustering identifies the typical normal data across different clusters, and manifold modeling attempts to capture the underlying structure of these normal data by reconstruction, thus identifying those typical noisy data by the reconstruction error. Finally, explicit supervision combines the typical normal and noisy data to train a discriminator, achieving a more robust and accurate detection result.}
\label{fig3}   
\end{figure*}

\section{Methodology}\label{sec:Methodology}

\subsection{Model overview}
An overview of our Diff-CXR is shown in \textcolor{newcolor}{Fig. \ref{fig2}}. Firstly, the latent noise filtering strategy (LNFS) projects all images into the latent space where visual-relevant details are well preserved, and efficiently removes those data anomalies with their reports in such latent space, following a coarse-to-fine fashion. Then, within the curated dataset, the adaptive vision-aware textual learning strategy prompts the domain-specific language model to learn essential and succinct report embeddings, providing textual guidance for generation. Finally, disease-knowledge enhanced diffusion-based TTI model is trained in two stages, including vanilla diffusion process where only textual condition is injected, and disease-specific knowledge guidance to refine generation results gradually. 
\subsection{Latent noise filtering strategy}
\label{sec:3.2}
Inspired by the representative latent space where diffusion process is performed, here, all images are encoded by the pretrained autoencoder $\boldsymbol{\varepsilon}$, and LNFS is devised to gradually learn the distribution discrepancy between normal and noisy images in the latent space through three stages, including latent clustering, manifold modeling and explicit supervision. The framework of LNFS is illustrated in \textcolor{newcolor}{Fig. \ref{fig3}}.

{\bfseries Pretrained autoencoder}: 
In this module, the objective is to construct a latent space of input CXRs, where perceptually-related attributes are well preserved. It has been proven that the autoencoder component of SD can precisely reconstruct CXRs out-of-box \cite{PierreFewshotSD2022, RoentGen2022}. Consequently, we keep the autoencoder component frozen to derive the latent representations of input images. Assume that the training set $\{x_k,y_k\}_{k=1}^N$ contains \textit{N} (image, report) pairs. The encoder $\boldsymbol{\varepsilon}$ converts $\{x_k\}_{k=1}^N$ into $\{z_k\}_{k=1}^N$, and the decoder ${D}$ can reconstruct images from the latent space, giving $x'=D(z_k)$. Both data curating and diffusion process are performed in such latent space.

{\bfseries Latent clustering}: 
Considering that humans learn the ability to recognize anomalies based on thousands of non-anomalous samples they have known, the first stage of LNFS is K-means clustering process behaving as the preliminary step to identify the major part of normal images. Detailed in \textcolor{newcolor}{Fig. \ref{fig3}}, latent representations are reshaped into a one-dimensional vector $\{v_k\}_{k=1}^N$, and then K-means clustering is directly applied to partition these vectors $\{v_k\}_{k=1}^N$ into different clusters, which are denoted by different centroids $\{c_k\}_{k=1}^m$, where \textit{m} is the numbers of centroids. All images are sorted based on euclidean distances between their latent representations $v_k$ and corresponding centroids $c_k$ in the ascending order. By a predefined distance threshold, the typical normal data can be filtered out based on the assumption that anomalies typically manifest in locations distant from their nearest neighbors, whereas normal instances reside within densely clustered regions. 

{\bfseries Manifold modeling}: Given the typical normal data filtered from latent clustering, manifold modeling aims to understand the general distribution of non-anomalous data $\{z_k\}_{k=1}^N$ in the latent space and captures the underlying structure of normal data. In this stage, a U-shape Autoencoder, composed of an encoding network and a decoding network, is devised to reconstruct the normal instances in the latent space by Mean Square Error (MSE). Through reconstruction, the U-shape Autoencoder is forced to retain the signification information which is relevant to the normal data, so those anomalies that differ from the major normal instances should be poorly reconstructed. The reconstruction error is directly used as anomaly score to evaluate those suspicious data excluded from the clustering process, thus identifying those typical noisy data. 

{\bfseries Explicit supervision}: Given the typical normal and noisy data, a latent noise discriminator is designed to explicitly model both noisy and normal samples. As shown in \textcolor{newcolor}{Fig. \ref{fig3}}, the discriminator consists of a sequence of cascaded convolutional layers, normalization layers, and activation layers for hierarchical feature extraction, followed by an adaptive average pooling layer and a linear classifier for binary classification. The discriminator is supervised under the binary cross entropy loss, so the output score that determines whether the input is noisy pattern can also be employed as anomaly score. 

Finally, all noisy images are detected by the combination of the reconstruction error and discrimination score. The clean (CXR, report) pairs can be denoted as: $\{x_k,y_k\}_{k=1}^n$, where \textit{n} represents the number of entries in the pruned dataset.

\subsection{Adaptive vision-aware textual learning strategy}
Textual learning is conducted with clean (report, CXR) pairs $\{x_k,y_k\}_{k=1}^n$. To obtain the informative and concise report embeddings, here we design an adaptive vision-aware textual learning strategy to sufficiently explore the report comprehension capabilities inherent in the large domain-specific VLM, ensuring that those visually relevant features can be well extracted and preserved. 

\begin{figure}[!t]
\centerline{\includegraphics[width=\columnwidth]{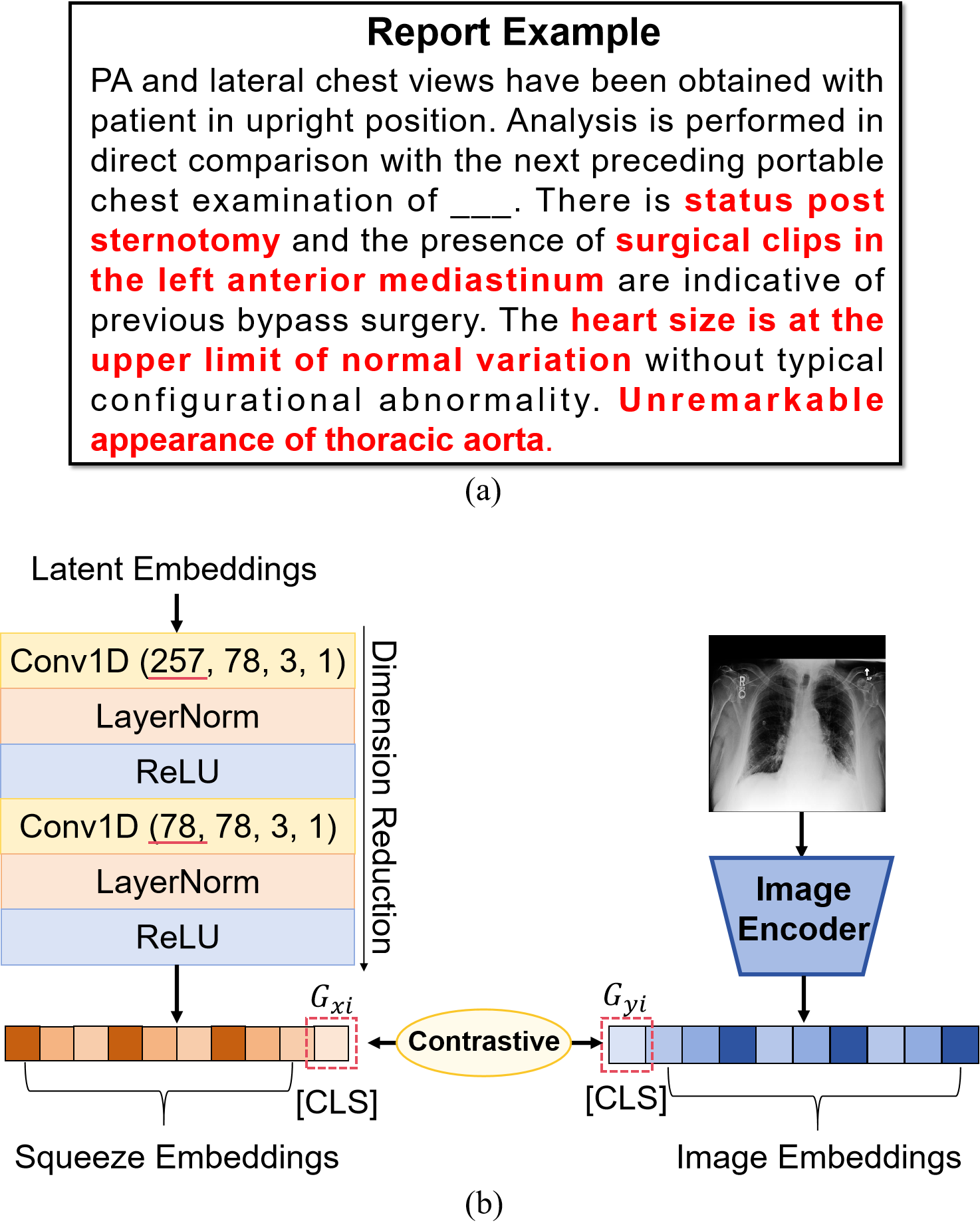}}
\caption{Illustration of AVA-TLS. (a) presents a medical report example, where CXR-related tokens are highlighted. The report should also be padded by useless tokens to meet the default token limits during tokenization. (b) denotes the structure of our information squeeze module.}
\label{fig4}
\end{figure}

As shown in \textcolor{newcolor}{Fig. \ref{fig4}a}, for a long medical report, there are a certain number of image-unrelated tokens in the report, which are useless for our TTI generation. To capture concise image-related context, an effective information squeeze module is introduced to forcefully compress latent embeddings into a lower dimension via a contrastive objective. Detailed in \textcolor{newcolor}{Fig. \ref{fig4}b}, the module consists of two cascaded blocks. Each block contains one-dimensional convolutional layer followed by the layer normalization and rectified linear unit (ReLU) activation functions. The information squeeze module is inserted at the end of the VLM text encoder, which outputs squeezed latent embeddings for textual guidance. To preserve those visual-concerned textual features during compression, the symmetric contrastive loss is introduced to apply vision supervision to report comprehension. Both image and text encoders of BiomedCLIP are employed and initialized with pretrained weights to map both modalities to specific latent space. Specifically, the contrastive objective jointly trains the image encoder and text encoder by maximizing the cosine similarity of the image and text embeddings of \textit{T} real pairs in the batch while minimizing the cosine similarity of those false pairs in the batch. Notably, the contrastive objective optimizes for text-image retrieval, so those image-unrelated report parts will be weakened or ignored during the alignment process, which is well suited for our purpose to extract crucial text embeddings from medical reports.

The projected [CLS] tokens are used as the global representations, which can be denoted as ($G_{xi}$,$G_{yi}$). $xi$, $yi$ representes $i_{th}$ CXR and its corresponding report. InfoNCE loss is adopted as the contrastive objective, which can be detailed as follows:

\begin{equation}
L(G_{xi}, G_{yi})=-\log(\frac{\exp{\frac{G_{xi}^TG_{yi}}{\tau}}}{\sum_{k=1}^{T}\exp{\frac{G_{xi}^TG_{yk}}{\tau}}})
\end{equation}
\begin{equation}
L_{infoNCE}=\frac{1}{T}\sum_{i=1}^{T}[l(G_{xi}, G_{yi}) + l(G_{yi}, G_{xi})]
\end{equation}

Combined with the symmetric contrastive loss, the information squeeze module can explicitly model the context relationship between different latent tokens and dynamically aggregate important token information for textual guidance. 

\subsection{Disease-knowledge enhanced TTI generation}

To generate high-quality medical images from medical reports, we propose a novel disease-knowledge enhanced diffusion-based TTI model based on the clean data pruned by LNFS and textual guidance provided by AVA-TLS. The disease knowledge injection mechanism is designed to distill specific disease knowledge into the vanilla diffusion process, highlighting those disease-related parts in report embeddings and gradually refining the generation process. Next, we will introduce the basic formulation and the component details of our TTI model.

{\bfseries Basic formulation}: Latent diffusion models take the diffusion process in a more efficient and lower-dimensional latent space of powerful pretrained autoencoders. The noise injection process, termed as the forward process, models a fixed Markov chain of length T, which can be defined as:

\begin{equation}
q(z_{1:T}|z_{0})=\prod_{1}^{T}q(z_{t}|z_{t-1})
\end{equation}

\begin{equation}
q(z_{t}|z_{t-1})=N(z_{t}|\sqrt{1-\beta_{t}}z_{t-1}, \beta_{t}I)
\end{equation}

Here, ${z_{0}}$ is the latent representation of the clean sample. The noise depends on the time-varying variance $\beta_{t}\in(0,1)$, where $t\in{1,...T}$.
The learning task is to reverse this forward process by deducing ${z_{t-1}}$ from its corrupted version ${z_{t}}$ for $t\leq T$. The corresponding objective can be expressed as:

For our disease-knowledge enhanced diffusion model, conditioning information including the report embeddings and disease knowledge embeddings is further fed into vanilla diffusion process, the noise prediction object can be detailed as:

\begin{figure*}[!t]
\centerline{\includegraphics[width=1.8\columnwidth]{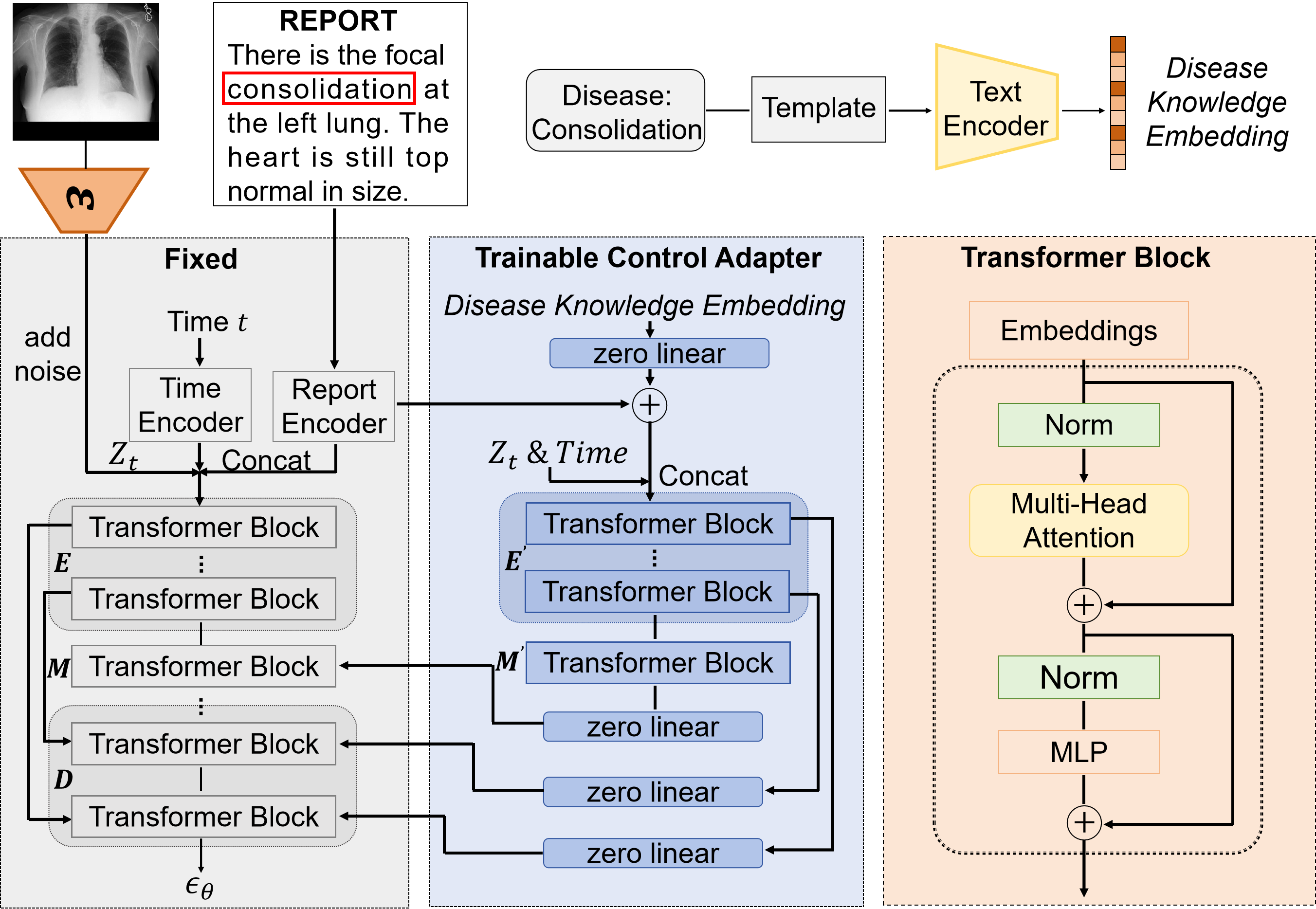}}
\caption{The illustration of control adapter. Original weights of the denoising model are fixed. The trainable control adapter, which consists of the encoder block and middle block copies, takes the disease knowledge embedding as input and interfaces with the fixed denoising model with the zero linear layers.}
\label{fig5}
\end{figure*}

\begin{equation}
\min_{\theta}E_{t,z_{0},c,\epsilon}=\| \epsilon - \epsilon_{\theta}(z_{t},t,c) \|_{2}^2
\end{equation}
where $\epsilon_{\theta}(z_{t},t,c)$ denotes a sequence of denoising autoencoders parameterized with $\theta$, which shares equal weights to predict the noise $\epsilon$. $c$ is the report condition or alternative condition information.

{\bfseries Vanilla denoising process}: The denoising model receives the noisy input $z_{t}$, diffusion timestep $t$ and conditioned report embeddings as input to predict injected noise as timestep $t$. The U-ViT architecture is adopted as the backbone of denoising model because it is capable of handling inputs from different modalities without introducing extra modules and takes fewer parameters than Stable Diffusion, greatly facilitating the report-to-CXR generation process.

{\bfseries Disease knowledge injection mechanism}: Medical reports typically provide highly detailed descriptions of disease conditions. Hence, it is crucial to maintain or enhance disease representations during the generation process for the accuracy and authenticity of generated images.

To this end, DKIM is introduced to incorporate general and more detailed disease knowledge guidance into the pretrained TTI model, refining and highlighting disease representations in the diffusion process. Disease types are generally ascertainable or predictable through provided clinical presentations in reports. Here, CheXbert \cite{chexbert} is utilized to annotate data with disease categories. A simple sentence template “There is \{disease type\}.” is employed to denote respective disease and “There is no finding.” is utilized to denote the normal condition without any disease symptoms. When multiple diseases are present in a single image, different disease templates will be concatenated to represent the overall health condition. Given that AVA-TLS can forge meaningful connections between related sections of reports and their corresponding disease representations in images, the pretrained text encoder is introduced to extract disease knowledge embeddings based on predefined disease templates to gain finer control of the report-to-CXR generation process.

As shown in \textcolor{newcolor}{Fig. \ref{fig5}}, a control adapter is proposed to gradually distill the disease knowledge into the diffusion process. Specifically, the weights of the vanilla denoising structure are fixed, and the encoder block, middle block, and decoder block are denoted as $E$, $M$, $D$, respectively. The encoding block and middle block are cloned to a trainable copy, designated as $E'$ and $M'$. $e (e')$, $m (m')$, and $d$ represent the output of certain encoder block, middle block or decoder block. Thereafter, the disease knowledge is incorporated from the control adapter during the decoding process. The input of $i_{th}$ block of the decoder block $D$ can be expressed as:
\begin{equation}
\begin{cases}
(m+zero(m'))\!+\!e_{j}\!+\!zero(e'_{j}) \quad i=1, i\!+\!j=9 \\
d_{i-1}\!+\!e_{j}\!+\!zero(e'_{j}) \quad 2\leq i \leq 8, i\!+\!j=9
\end{cases}
\end{equation}
where zero represents a zero linear year which weight and bias parameters are initialized to zero when the training starts. In this case, both $zero(:)$ equals to zero when the training starts, and the input of $i_{th}$ block of the decoder block $D$ can be re-expressed as:
\begin{equation}
\begin{cases}
m+e_{j} \quad i=1, i+j=9 \\
d_{i-1}+e_{j} \quad 2\leq i \leq 8, i+j=9
\end{cases}
\end{equation}
which is the same as the original denoising model, thus maintaining the generation capabilities of the pre-trained model. Therefore, zero-initialized layers effectively ensure that harmful noise will not influence the hidden states of network when the training starts.

Generally, the control adapter employs a zero linear layer to progressively modify the report representations on the context level based on given disease knowledge embeddings. Then, the trainable copy takes the modified report embeddings as input, and gradually distills the disease knowledge into the diffusion process to refine the generation results. 

\section{Experiments}\label{sec:Experiments}
\subsection{Materials and preprocessing}
Experiments are conducted on two widely used report-to-CXR generation benchmarks, \emph{i.e.}, MIMIC-CXR \cite{mimiccxr} and IU-Xray \cite{IUxray}. 

MIMIC-CXR is a large-scale (report, CXR) dataset, provided by the Beth Israel Deaconess Medical Center. The dataset includes 377,110 chest X-ray images and 227,835 reports. Each image is annotated with multiple classes of 14 diagnostic labels, and 11 common diseases are considered for benchmarking purposes. For a fair comparison, the official data splits are adopted.

IU-Xray from Indiana University is a relatively small dataset, which contains 7,470 chest X-ray images and 3,955 radiology reports. IU-Xray comes from the different hospital and its reports are more concise and standardized. By CheXbert labeler \cite{chexbert}, each image is also labeled with 14 diagnostic labels, and 11 diseases are utilized for evaluation. 

Only anterior-posterior (AP) / posterior-anterior (PA) scans are considered, and all images are resized to 256$\times$256. Following the previous work \cite{llmcxr2024}, we use MIMIC-CXR to develop our Diff-CXR model, and then validate on the test set of MIMIC-CXR. The whole IU-Xray dataset is further employed to gauge the performance of Diff-CXR towards unseen text.

To further evaluate the effectiveness of our approach, we conduct a series of downstream task experiments. Specifically, we use the MIMIC-CXR, ChestX-ray14 \cite{wang2017chestx}, and CheXpert \cite{irvin2019chexpert} datasets to validate our method on the multi-label thorax disease classification task.

ChestX-ray 14 contains 112,120 frontal-view X-rays from 30,805 patients with 14 disease labels. We follow the official data split which assigns 75,312 images for training and 25,596 images for testing. Images are resized from 1024$\times$1024 into 224$\times$224.

CheXpert is a large scale dataset consisting of 191,208 frontal-view chest X-rays. Within the dataset, images annotate 14 diseases. For evaluation, five common diseases, including atelectasis, cardiomegaly, consolidation, edema, and pleural effusion, are considered for benchmarking purposes. Images are resized to 224$\times$224 pixels and the evaluation is conducted on the official validation set.

\begin{table*}[h]
\caption{Implementation details of the components of Diff-CXR, which are trained separately.}
\label{tab}
\setlength{\tabcolsep}{12pt}
\centering
\begin{tabular*}{0.8\textwidth}{@{\extracolsep{\fill}}>{\centering\arraybackslash}p{80pt}> {\centering\arraybackslash}p{50pt}> {\centering\arraybackslash}p{50pt}> {\centering\arraybackslash}p{80pt}}
\hline
- & LNFS & AVA-TLS & Vanilla Diffusion\\
\hline
GPUs & 1A100 & 1A100 & 4A100\\
Batch size & 512 & 128 & 512 \\
Max training iters & 3.6k & 36k & 1M\\
Optimizer & Adam & AdamW & AdamW \\
Learning rate & 3e-5 & 1e-5 & 2e-4\\
Weight decay & 0 & 0 & 0.03\\
Betas & (0.9, 0.999) & (0.9, 0.999) & (0.9, 0.9)\\
Adjusted schedule & CosineAnnealingLR & CosineAnnealingLR & -\\
\hline
\end{tabular*}
\label{tab1}
\end{table*}

Moreover, we apply the MIMIC-CXR dataset to a cross-modal CXR-report generation task, demonstrating the flexibility and versatility of our proposed method.

\subsection{Evaluation measures}
\label{sec:4.2}
To gauge the quality of generated data, we utilize Fréchet Inception Distance (FID), which quantifies the distribution discrepancy between generated data and real data. Lower FID score indicates better vision fidelity. 

To assess vision-language alignment between generated images and input reports, we calculate the Mean Area Under the Curve (mAUC) score of generated data, against the original CheXbert labels from MIMIC-CXR. A pretrained thorax disease classification network, densnet121-resnet224-all \cite{huang2017densely, torchxrayvision}, is utilized to predict generated images. Higher mAUC score indicates better vision-language alignment between input reports and generated CXRs. 

Floating Point Operations (FLOPs) are employed to evaluate the computations complexity of different methods. 

In downstream multi-label thorax disease classification tasks, mAUC on different classes is reported for comparison and higher mAUC score means superior classification performance. For CXR-report generation, we employ the widely-used natural language generation (NLG) metrics and adopt the standard evaluation protocol to calculate the captioning metrics: BLEU \cite{bleu}, METEOR \cite{Meteor} and ROUGE-L \cite{Rouge}. Higher BLEU, METEOR, and ROUGE-L scores indicate a greater similarity between generated reports and reference reports.

\subsection{Experimental setting}
All components of the proposed method are implemented with PyTorch and are trained with NVIDIA Graphics Device A100 80GB GPUs, separately. Implementation Details are listed in \textcolor{newcolor}{Table \ref{tab1}}.

For LNFS, to train the reconstruction network and the discriminator, the Adam optimizer was used to update parameters with a batch size 512. The learning rate was set to 3e-5 and was adaptively adjusted based on the CosineAnnealingLR. 

For AVA-TLS, we adopt BiomedCLIP \cite{BiomedClip2023} as our baseline model. The AdamW optimizer was used in training with a batch size 128. The learning rate was set to 1e-5 and was also adaptively adjusted based on the CosineAnnealingLR. 

The disease-knowledge guided TTI model is trained in two stages. First, vanilla diffusion process is performed. AdamW optimizer was employed with a batch size 512, using a learning rate of 2e-4, to update parameters. Second, the pretrained TTI model is kept frozen, disease knowledge is embedded, and only control adapter is trained for additional 200K steps. Training parameters are consistent with the first phase.

\section{Result}\label{sec:Result}
\subsection{Ablation study}

\begin{table}
\caption{Ablation experiments on the three key components, LNFS, AVA-TLS, and DKIM}
\label{table}
\setlength{\tabcolsep}{12pt}
\centering
\begin{tabular}{c c c c c}
\hline
LNFS & AVA-TLS & DKIM & FID & mAUC\\
\hline
- & - & - & 28.84 & 0.644\\
\checkmark & - & - & 20.93 & 0.670 \\
\checkmark & \checkmark & - & 21.56 & 0.676\\
\checkmark & \checkmark & \checkmark & \textbf{19.50} & \textbf{0.678} \\
\hline
\end{tabular}
\label{tab2}
\end{table}

We separately ablate the LNFS, AVA-TLS, and DKIM to corroborate their contributions to report-to-CXR generation performance.

\subsubsection{The necessity of the latent noise filtering strategy}
\begin{figure}[!h]
\centerline{\includegraphics[width=\columnwidth]{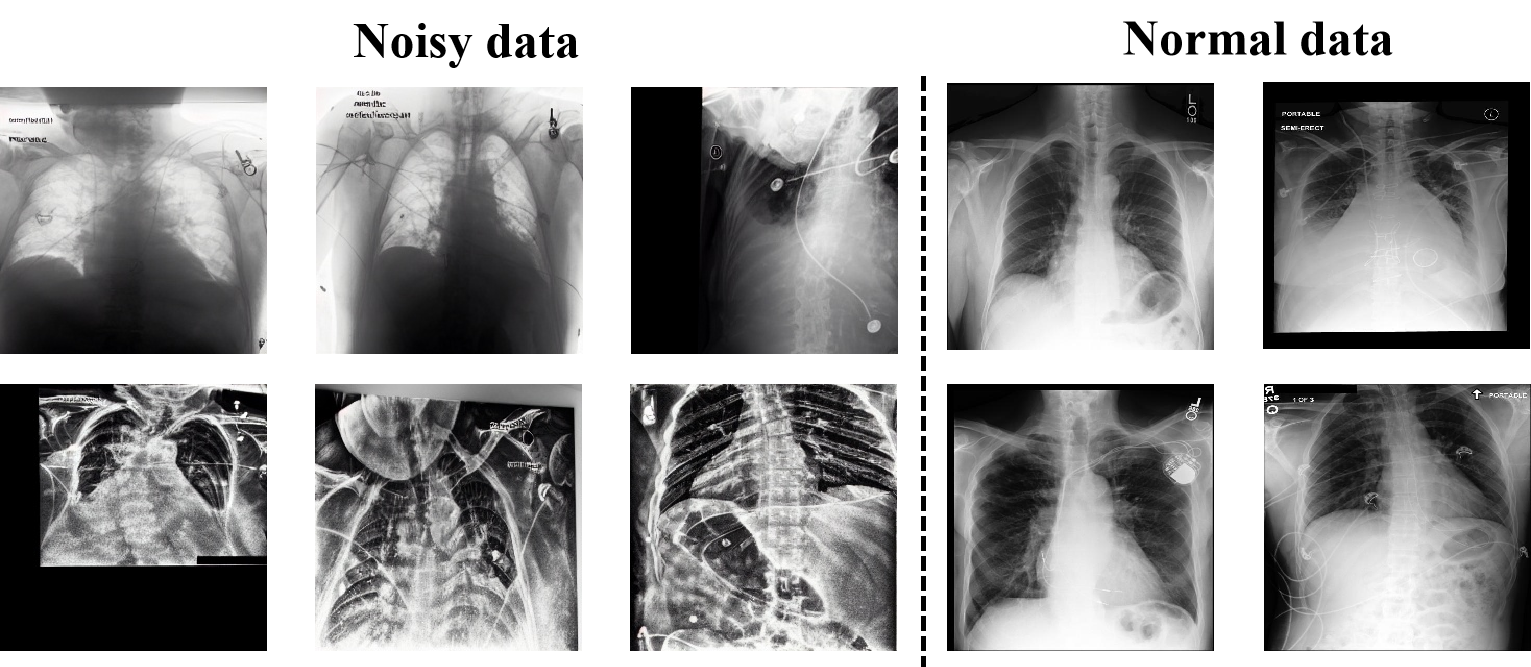}}
\caption{The visualization of filtered noisy data.}
\label{fig6}
\end{figure}

As illustrated in \textcolor{newcolor}{Fig. \ref{fig6}}, we visualize multiple noisy images filtered by LNFS and provide several normal images as a reference. We can observe that these abnormal samples exhibit notable deviations from normal images, featuring low signal-to-noise ratio and shape distortion, et al. These deviations also result in the lack of the alignment with their respective medical reports. 

\begin{figure}[!h]
\centerline{\includegraphics[width=\columnwidth]{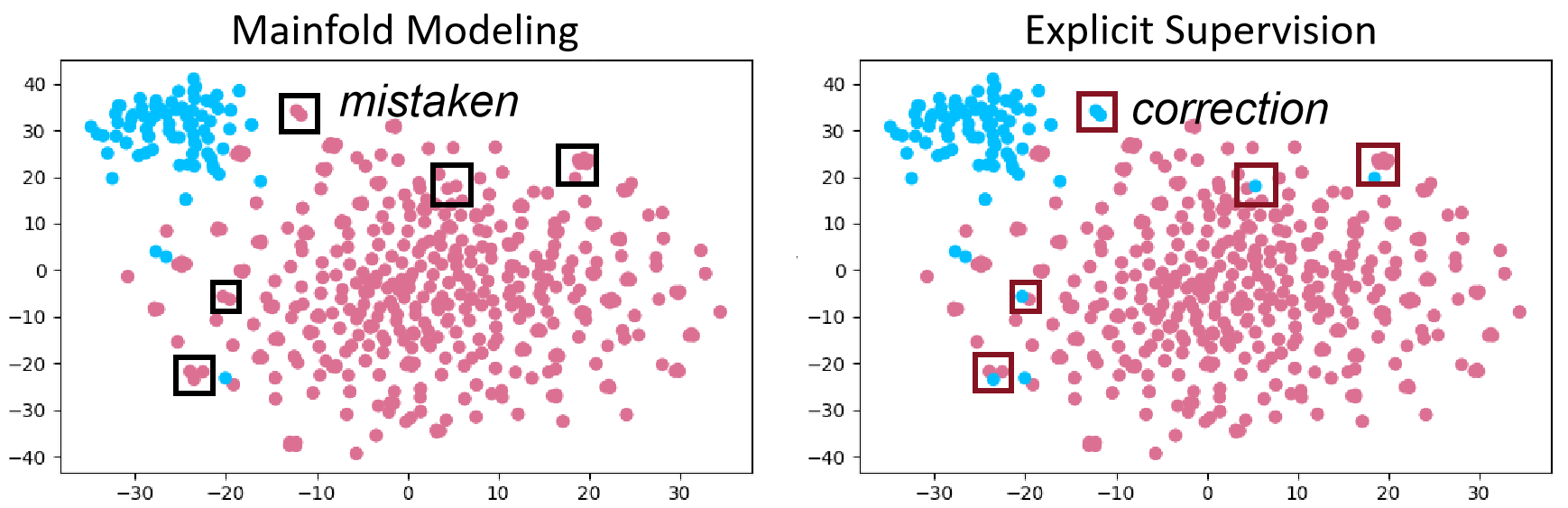}}
\caption{TSNE visualization of filtered noisy data. Blue points denote noisy data while red points denote normal data.}
\label{fig7}
\end{figure}

To demonstrate the superiority of our coarse-to-fine anomaly detection method, we compare the filtered samples generated by manifold modeling and explicit supervision phases using TSNE. As shown in \textcolor{newcolor}{Fig \ref{fig7}}, several noisy points near the decision boundary or nearby normal points, are misclassified as normal data during the manifold modeling stage. However, explicit supervision successfully identifies such noisy points and corrects the mistakes. Finally, 2000 images are identified as noisy data and excluded from training.

Finally, we compare the performance of Diff-CXR with and without LNFS on MIMIC-CXR. As shown in \textcolor{newcolor}{Table \ref{tab2}}, the overall performance of Diff-CXR with LNFS is better than Diff-CXR without LNFS. The FID score decreases from 28.84 to 20.93 and the mAUC score increases from 0.644 to 0.670, proving the significant impact of noisy images on generation process.

\subsubsection{The efficiency and precision improvements brought by the adaptive vision-aware textual learning strategy}

\begin{table}
\caption{Ablation study of AVA-TLS on squeezed token length.}
\label{table}
\setlength{\tabcolsep}{13pt}
\centering
\begin{tabular}{c c c c}
\hline
Token Length & FID & mAUC & GFLOPs \\
\hline
256 & - & - & 29.729 \\
128 & 21.25 & 0.675 & 22.287 \\
77 & 21.56 & \textbf{0.676} & 19.322\\
32 & \textbf{20.69} & 0.672 & \textbf{16.075} \\
\hline
\end{tabular}
\label{tab3}
\end{table}

According to the quantitative evaluation results shown in \textcolor{newcolor}{Table \ref{tab2}}, AVA-TLS improves the mAUC score from 0.670 to 0.676, with a slight decrease in FID score. Furthermore, as illustrated in \textcolor{newcolor}{Table \ref{tab3}}, when token length is set as 77, AVA-TLS substantially reduces the computational complexity to 19.322 GFLOPs, which is nearly two-thirds of the original complexity, 29.729 GFLOPs. In general, AVA-TLS enables Diff-CXR to achieve a trade-off between the model complexity, vision-language alignment and authenticity of generated images successfully.

Next, we compare the performance of AVA-TLS on different token length after compression. Obviously, when the token length decreases from 256 to 32, the complexity decreases from 29.729 GFLOPs to 16.705 GFLOPs. As shown in \textcolor{newcolor}{Table \ref{tab3}}, token 77 achieves the best mAUC score while token 32 achieves the best FID score. We assume that reduced tokens lead to smaller distribution discrepancies between different textual embeddings, so that model can easily reference corresponding images through fewer textual tokens. However, there may be some textual semantic information loss during the compression, further influencing the vision-language alignment of generated results. In following experiments, we adopt token 77 as our best configuration, because the consistency between generated images and input reports is considered the most crucial aspect in medical image generation tasks.

\subsubsection{The image detail refinement benefiting from the disease-knowledge injection mechanism}
\begin{figure}[!h]
\centerline{\includegraphics[width=\columnwidth]{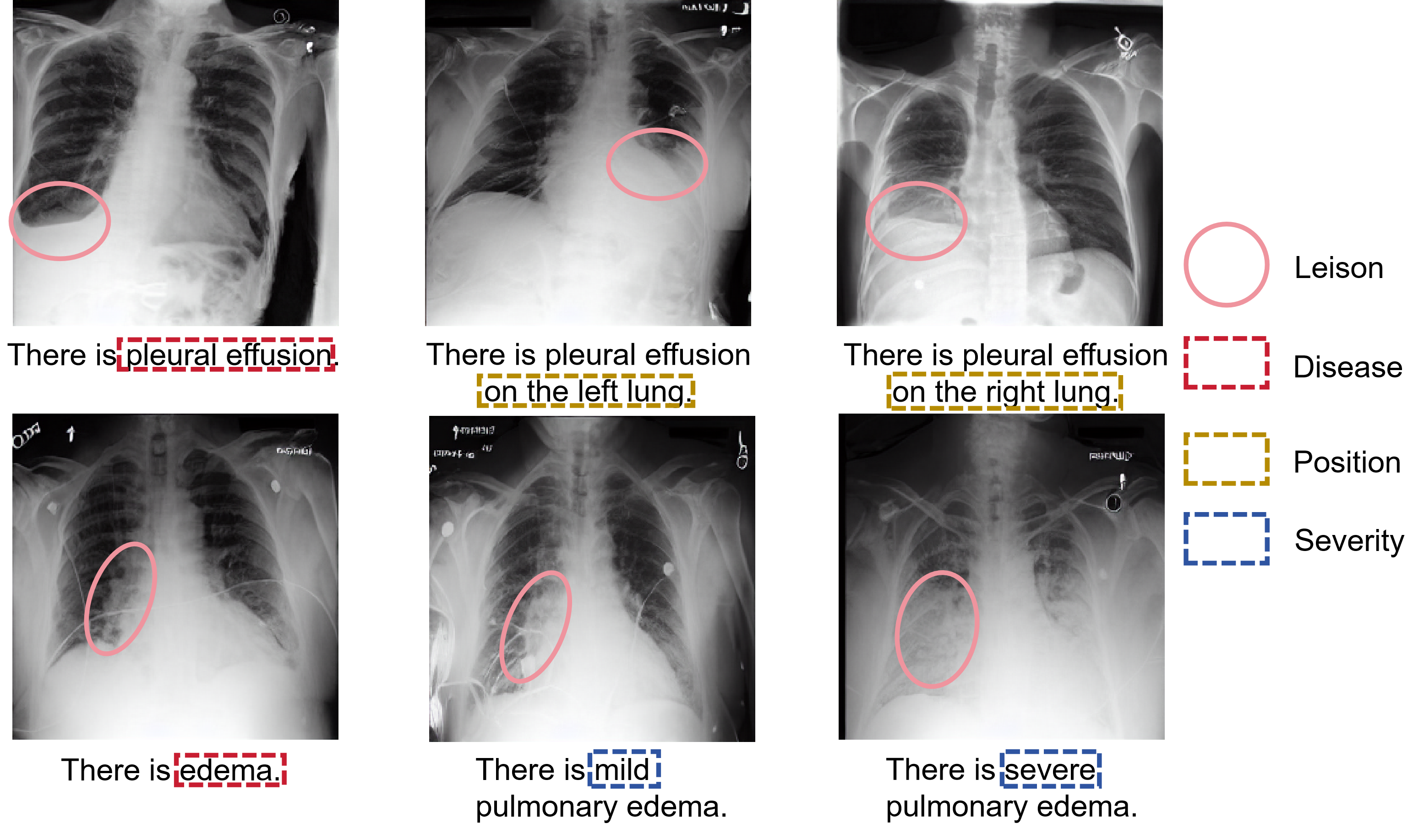}}
\caption{Illustration of generated CXRs based on given disease template sentences.}
\label{fig8}
\end{figure}

Firstly, to demonstrate that the feature embeddings obtained by the Vision Language Model (VLM) carry accurate disease information for a specific disease template sentence, we visualize CXRs generated from several disease template examples. As shown in \textcolor{newcolor}{Fig. \ref{fig8}}, Diff-CXR can successfully synthesize CXRs conditioned on the disease template sentence encoded by the domain-specific VLM. In addition, with slight modifications on the template sentence, Diff-CXR can generate realistic CXRs with various lesions of different severity levels and locations. For example, Diff-CXR is capable of generating CXRs with pleural effusions in both left and right lungs, as well as varying degrees of edema, from mild to severe.

We perform relevant ablation experiments on the effectiveness of DKIM of our Diff-CXR, which are presented in \textcolor{newcolor}{Table \ref{tab2}}. As observed in the table, it is worth noting that DKIM decreases the FID value from 21.56 to 19.50 and increases the mAUC score from 0.676 to 0.678, demonstrating that DKIM successfully improves the realism and accuracy of generated CXRs.

\begin{table}[!h]
\vspace{25pt}
\centering
    \caption{Comparison results with other SOTA medical TTI methods on FID score and computation complexity.}\label{table}
    \setlength{\tabcolsep}{1pt}
    \begin{tabular}{p{45pt} p{50pt} p{50pt} p{50pt} p{42pt}}
    \hline
    \centering
     & FID \par (MIMIC) & FID \par (IU-Xray) & Complexity \par (GFLOPs) & Inference \par time (s) \\
    \hline
    UnixGen & 97.861 & 79.690 & 90.460 & 27.455\\
    LLM-CXR & 29.270 & 41.288 & 275.453 & 9.780\\
    RoentGen & 56.499 & 73.316 & 86.307 & 5.185 \\
    Diff-CXR & \textbf{19.500} & \textbf{31.459} & \textbf{29.641} & \textbf{1.295} \\
    \hline
    \end{tabular}\label{tab4}
    \vspace{-40pt}
\end{table}

\subsection{Comparison with SOTA medical TTI methods}
\label{sec:5.2}

\begin{table*}[!h]
\centering
\caption{Comparative analysis of vision-language alignment of generated results for SOTA medical TTI methods using mAUC score.}
\label{table}
\begin{threeparttable}  
\setlength{\tabcolsep}{5pt}
\begin{tabular*}{1.0\textwidth}{@{\extracolsep{\fill}}>{\centering\arraybackslash}p{50pt}> {\centering\arraybackslash}p{20pt}> {\centering\arraybackslash}p{20pt}> {\centering\arraybackslash}p{20pt}> {\centering\arraybackslash}p{20pt}> {\centering\arraybackslash}p{20pt}> {\centering\arraybackslash}p{20pt}> {\centering\arraybackslash}p{20pt}> {\centering\arraybackslash}p{20pt}> {\centering\arraybackslash}p{20pt}> {\centering\arraybackslash}p{20pt}> {\centering\arraybackslash}p{20pt}> {\centering\arraybackslash}p{20pt}}
\hline
 MIMIC & Atel. & Cnsl. & Pmx. & Ede. & Eff. & Pna. & Cml. & Les. & Frc. & Opc. & ECm. & Avg.\\
\hline
UnixGen & 0.495 & 0.494 & 0.463 & 0.503 & 0.496 & 0.505 &	0.485 &	0.486 &	0.522 &	0.474 &	0.472 &	0.491\\
LLM-CXR & 0.662 & 0.602 & 0.722 & 0.754 & 0.781 & 0.546 &	0.661 &	0.477 &	0.482 &	0.640 &	0.581 &	0.628\\
RoentGen & 0.590 & 0.505 & 0.523 & 0.602 & 0.599 & 0.518 & 0.648	& 0.474	& \textbf{0.574} & 0.554 & 0.530 & 0.556\\
Diff-CXR & \textbf{0.693} & \textbf{0.652} & \textbf{0.766} & \textbf{0.808} & \textbf{0.840} & \textbf{0.571} & \textbf{0.737} & \textbf{0.561} & 0.546 & \textbf{0.651} & \textbf{0.628} & \textbf{0.678}\\
\hline
IU-Xray & Atel. & Cnsl. & Pmx. & Ede. & Eff. & Pna. & Cml. & Les. & Frc. & Opc. & ECm. & Avg.\\
\hline
UnixGen & 0.511 & 0.505 & 0.517 & 0.512 & 0.530 &	0.517 &	0.515 &	0.428 &	0.501 &	0.507 &	0.432 &	0.498 \\
LLM-CXR & 0.525 & 0.497 & 0.472 & 0.508 & 0.458 &	0.519 &	0.501 &	0.466 &	0.419 & 0.513 &	0.495 &	0.488 \\
RoentGen & 0.522 & 0.522 & 0.515 & 0.459 & 0.537 & 0.475 & 0.602 & 0.602 & 0.550& 0.532 & 0.563 & 0.534 \\
Diff-CXR & \textbf{0.754} & \textbf{0.831} & \textbf{0.614} & \textbf{0.577} & \textbf{0.943} & \textbf{0.550} & \textbf{0.835} & \textbf{0.769} & \textbf{0.827} & \textbf{0.796} & \textbf{0.903} & \textbf{0.763} \\
\hline
\end{tabular*}
\begin{tablenotes}
\footnotesize 
\item Atel.: Atelectasis, Cnsl.: Consolidation, Pmx.: Pneumothorax, Ede.: Edema, Eff.: Pleural Effusion, Pna.: Pneumonia, Cml.: Cardiomegaly, Les.: Lung Lesion, Frc.: Fracture, Opc.: Lung Opacity, ECm.: Enlarged Cardiomediastinum, Avg.: Average.
\end{tablenotes}
\end{threeparttable}
\label{tab5}
\end{table*}

Prevalent medical TTI methods can be categorized into two types: autoregressive model and diffusion model. For autoregressive-based TTI methods, we choose UnixGen \cite{lee2024vision} 
and LLM-CXR \cite{llmcxr2024} for comparison while for diffusion-based TTI methods, RoentGen \cite{RoentGen2022} is employed.

\begin{figure*}[!t]
\centerline{\includegraphics[width=1.0\textwidth]{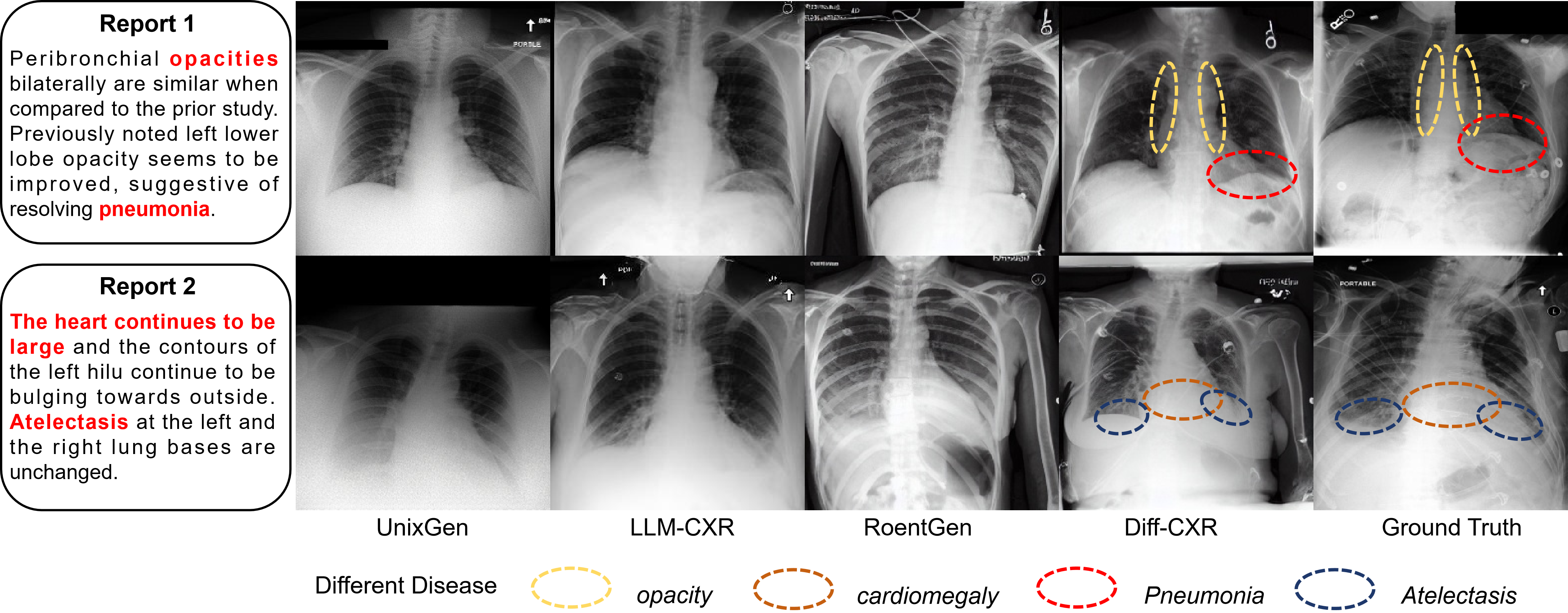}}
\caption{Visual comparisons between different methods. Disease parts are highlighted in red and corresponding disease imaging characteristics in Diff-CXR and ground truth are cycled in different colors.}
\label{fig9}
\end{figure*}

Experiments are conducted on MIMIC-CXR and IU-Xray to test both in-of-distribution and out-of-distribution performance. As shown in \textcolor{newcolor}{Table \ref{tab5}}, the AUC scores of images generated by Diff-CXR for nearly 11 common thorax diseases are consistently higher than those of LLM-CXR, indicating the best alignment between generated images and input reports. From \textcolor{newcolor}{Table \ref{tab4}}, our Diff-CXR also surpasses LLM-CXR on the FID score by 9.770 and 9.829, respectively, proving the superiority of Diff-CXR in vision fidelity. As shown in \textcolor{newcolor}{Fig. \ref{fig9}}, related disease representations are highlighted in images generated by Diff-CXR and the ground truth. Obviously, images generated by Diff-CXR share the similar disease imaging characteristics with the ground truth, underscoring the superiority of our method. Furthermore, our Diff-CXR demonstrates excellent generalizability towards unseen text with different distributions. Compared to LLM-CXR in IU-Xray, our Diff-CXR consistently delivers commendable results with an improvement of 56.4\% and 23.8\% in mAUC and FID score, respectively.  

As indicated in \textcolor{newcolor}{Table \ref{tab4}}, our Diff-CXR also achieves the lowest computation complexity 29.641 GFLOPs, nearly one-third that of RoentGen and one-ninth that of LLM-CXR, successfully demonstrating the efficiency of our method. In addition, the inference speed of our Diff-CXR is significantly enhanced due to the optimization in computational complexity, which is about 7.6 times and 4.0 times faster than that of LLM-CXR and RoentGen.

\subsection{Experiments on downstream tasks}
As shown in \textcolor{newcolor}{Fig. \ref{fig8}}, synthetic images precisely reflect lesion types, positions and severity levels. Having such a powerful generative model enables us to perform data augmentation. Here, we synthesize novel images using reports which contains lesions coming from MIMIC-CXR and validate the efficacy of generated images on multi-label thorax disease classification tasks and a CXR-report generation task on MIMIC-CXR. In total, 95,787 images are generated for downstream tasks.

\begin{table}[h]
\caption{Thorax disease classification tasks, which are evaluated by the mAUC score.}
\label{table}
\setlength{\tabcolsep}{5.0pt}
\centering
\begin{tabular}{c c c c c c c}
\hline
Method & 1\% real data &  10\% real data & 100\% real data\\
\hline
\textbf{MIMIC-CXR} & & & \\
DenseNet & 0.632 & 0.727 & 0.780 \\
+synthetic & \textbf{0.752} & \textbf{0.764} & \textbf{0.784}\\
VIT \rule{0pt}{10pt} & 0.672 & 0.762 & 0.792 \\
+synthetic & \textbf{0.764} & \textbf{0.785} & \textbf{0.799}\\
ConvNext & 0.612 & 0.744 & 0.790\\
+synthetic  & \textbf{0.752} & \textbf{0.768} & \textbf{0.794}\\
maxVIT & 0.627 & 0.724 & 0.789\\
+synthetic  & \textbf{0.742} & \textbf{0.767} & \textbf{0.791}\\
\hline
\textbf{ChestX-ray14} & & & \\
DenseNet & 0.560 & 0.713 & 0.804 \\
+synthetic & \textbf{0.764} & \textbf{0.788} &\textbf{ 0.815}\\
VIT \rule{0pt}{10pt} & 0.594 & 0.753 & 0.820 \\
+synthetic & \textbf{0.780} & \textbf{0.804} & \textbf{0.824}\\
ConvNext & 0.575 & 0.725 & 0.807\\
+synthetic & \textbf{0.775} & \textbf{0.795} &\textbf{0.818}\\
maxVIT & 0.552 & 0.730 & 0.805\\
+synthetic & \textbf{0.777} & \textbf{0.793} & \textbf{0.821}\\
\hline
\textbf{ChestXpert} & & & \\
DenseNet & 0.775 & 0.841 & 0.871 \\
+synthetic & \textbf{0.859} & \textbf{0.878} & \textbf{0.884}\\
VIT \rule{0pt}{10pt} & 0.819 & 0.870 & 0.878 \\
+synthetic & \textbf{0.866} & \textbf{0.884} & \textbf{0.890}\\
ConvNext & 0.784 & 0.858 & 0.878\\
+synthetic & \textbf{0.866} & \textbf{0.877} & \textbf{0.889}\\
maxVIT & 0.778 & 0.855 & 0.883\\
+synthetic & \textbf{0.879} & \textbf{0.876} & \textbf{0.885}\\
\hline
\end{tabular}
\label{tab6}
\end{table}

For classification,  we firstly employ DenseNet-121 pre-trained on ImageNet, following CheXclusion \cite{ChexClusion}. Then, we compare ViT pre-trained on thousands of chest X-rays by MAE, which has been proven to be SOTA pretrained methods for thorax disease classification \cite{VITMAE}. To further demonstrate the generalizability of our method, we also test two commonly used classifiers in natural domains: ConvNext \cite{liu2022convnet} and maxVIT \cite{tu2022maxvit}. In \textcolor{newcolor}{Table \ref{tab6}}, it is obvious that our synthetic data consistently improves the performance of various models trained on different datasets using various ratios of real data (1\%, 10\%, 100\%). Notably, with only 1\% real data, synthetic data can potentially improve the model's mAUC metric ranging from 0.047 to 0.225 across all datasets. Especially, the mAUC value of maxVIT is substantially enhanced by as much as 40.76\% on ChestX-ray 14, fully demonstrating the importance of our synthetic data in limited data scenarios.

For report generation, two radiology report generation methods, including R2GenCMN \cite{R2GENCMN} and XPRONET \cite{wang2022cross}, are finetuned on MIMIC-CXR and our synthetic (report, CXR) pairs based on their pretrained weights. The results, presented in \textcolor{newcolor}{Table \ref{tab7}}, reveal  a general improvement in report generation performance. Specifically, with Diff-CXR, XPRONET is capable of producing more realistic and accurate medical reports, demonstrated by an improvement of 0.009 and 0.006 in Bleu1 and Bleu2 scores, respectively. This demonstrates the realism and accuracy of our generated (report, CXR) pairs, confirming their efficacy as comparable to real images for training purposes. The simplicity and ability of our Diff-CXR to integrate with other models make for a noteworthy contribution.

\begin{table}[h]
\caption{Radiology Report Generation Task on MIMIC-CXR.}
\label{table}
\setlength{\tabcolsep}{2.0pt}
\centering
\begin{tabular}{c c c c c c c}
\hline
Method & Bleu1 & Bleu2 & Bleu3 & Bleu4 & Meteor & Rouge\_L \\
\hline
R2GenCMN & 0.353 & 0.218 & 0.145 & 0.103 & \textbf{0.142} & 0.277 \\
+synthetic & \textbf{0.358} & \textbf{0.224} & \textbf{0.151} & \textbf{0.108} & 0.141 & \textbf{0.280} \\
XPRONET \rule{0pt}{10pt} & 0.344 & 0.215 & 0.146 & 0.105 & 0.138 & \textbf{0.279} \\
+synthetic  & \textbf{0.353} & \textbf{0.221} & \textbf{0.150} & \textbf{0.108} & \textbf{0.142} & 0.278 \\
\hline
\end{tabular}
\label{tab7}
\end{table}

\section{Discussion}\label{sec:Discussion}
In this paper, we propose a unified learning framework for report-to-CXR generation through a disease knowledge-enhanced diffusion model. The performances are compared with the state-of-the-art methods on two widely used benchmarks and the results indicate that the proposed method outperforms the existing state-of-the-art methods. We will further discuss several important insights behind Diff-CXR.
\subsection{Towards realistic medical TTI generation}
Realistic images are essential for computer-aided diagnosis to ensure its reliability and effectiveness in clinical applications. However, in medical scenarios, diffusion-based TTI methods tend to overfit those noisy data, degrading the realism of generation results. As shown in \textcolor{newcolor}{Table \ref{tab2}}, Diff-CXR with LNFS outperforms Diff-CXR trained on the noisy dataset, yielding a remarkable 27.4\% improvement in FID and a 4.0\% increase in mAUC score. Here, we further delve into why those noisy samples seriously impede the Diffusion-based TTI generation performance. First, in the report-to-CXR generation setting, these noisy samples are treated as supervisory signals instead of the input signals. Once TTI model overfits these noisy patterns, the realism of generated CXRs will degrade seriously. Second, TTI task is data-hungry, while medical datasets with paired images and reports tend to be limited, which is far less than natural image-text datasets that can easily reach billion-scale. Thus, such noisy images result in fewer effective pairs of images and reports, leading to worse impacts on the generation task. Third, in the reverse diffusion process, the denoising model directly predicts the noise added onto the original input at arbitrary timestep $t \leq T$. In this case, the denoising model, which demonstrates strong learning capabilities, processes those noisy patterns for multiple times, further increasing the risk of overfitting.

\subsection{Towards efficient and accurate medical TTI generation}
The trade-off between efficiency and accuracy are crucial for the clinical implementation of medical TTI generation. For transformer-based autoregressive TTI methods, large parameter size and computational complexity limit their real applications. Diffusion models achieve comparative generation performance with a more reasonable model size. However, repeated evaluations on a noisy version of the input latent still make the training and inference of diffusion model expensive. With the token limits of medical VLM growing from 77 to 256, as depicted in \textcolor{newcolor}{Table \ref{tab3}}, the computation complexity of the denoising model increases from 19.322 GFLOPs to 29.727 GFLOPs. Therefore, in medical scenes, such training and inference costs are only expected to be higher. \textcolor{newcolor}{Fig. \ref{fig4}a} points that there are several image-unrelated tokens in the specific report. Additionally, we also conduct the statistics of token length on MIMIC-CXR, which shows that the length of report tokens primarily falls within the range of 50 to 150. Hence, for major reports, large domain-specific VLM will introduce several image-unrelated tokens into the hidden space. As textual guidance, these report embeddings may hinder the efficiency of overall report-to-CXR generation. In this case, our Diff-CXR employs AVA-TLS to prompt the large pretrained medical VLM to learn concise and crucial report embeddings for conditioned image generation. As presented in \textcolor{newcolor}{Table \ref{tab2}}, AVA-TLS enhances the mAUC score of generated images from 0.670 to 0.676. This improvement demonstrates that the effectiveness of AVA-TLS in extracting CXR-related report information to guide the image generation process and strengthen the alignment between input reports and generated CXRs. In addition, \textcolor{newcolor}{Fig. \ref{fig8}} illustrates that TTI model can generate accurate CXRs based on the disease-knowledge embeddings encoded by AVA-TLS, further proving that AVA-TLS is capable of accurately capturing disease representations in clinical reports. 

\subsection{Towards high-quality medical TTI generation}
For clinical application, medical field typically demands high-quality images, posing greater challenge for medical TTI task. Currently, SOTA medical TTI method mainly relies on tailored training strategies to enhance the quality of generated results. For example, LLM-CXR employs a two-stage fine-tuning strategy to gradually learn the intimate mapping between vision and language modalities. RoentGen extensively explores the fine-tuning strategies of Stable Diffusion to overcome the distribution shift between natural texts and medical reports.
However, the quality of their generated images is still limited due to the intrinsic characteristics of medical images and reports. Here, we focus on the impact of disease knowledge on medical TTI. Medical reports mainly consist of detailed descriptions of patients’ disease condition, so that the disease information is crucial for medical TTI. In this case, Diff-CXR designs DKIM to incorporate specific disease knowledge to optimize report representations and implement more refined control over the diffusion process, thus improving the synthesis quality. As shown in \textcolor{newcolor}{Table \ref{tab2}}, DKIM successfully improve the FID and mAUC score by 2.06 and 0.002, underscoring its effectiveness in enhancing the realism and accuracy of generated results.
High-quality medical TTI generation can serve as a more powerful data augmentation method for more diverse downstream tasks. \textcolor{newcolor}{Table \ref{tab6}} firstly illustrates that models trained on the combination of synthetic data and realistic data can achieve a remarkable improvement. Notably, with only 1\% realistic data and our synthetic data, the mAUC score of maxVIT achieves a significant improvement, which is 0.004 lower than maxVIT trained solely on 100\% real data, proving that our Diff-CXR can effectively address shortfalls of underrepresented conditions or subgroups caused by the data scarcity. In addition, images generated by Diff-CXR are not confined to traditional thoracic disease classification tasks. Detailed in \textcolor{newcolor}{Table \ref{tab7}}, they can also be effectively applied to cross-modal CXR-report generation tasks and successfully enhance the performance of typical report generation models. This also demonstrates the great importance of Diff-CXR for medical computer-aided diagnosis.

\subsection{Limitations}
Diff-CXR still has limitations. Currently, Diff-CXR only supports reports as conditional inputs, making it unsuitable for image segmentation tasks. Additionally, Diff-CXR has been trained exclusively on CXR datasets, which limits its applicability to other modalities and organs.

\section{Conclusion}\label{sec:Conclusion}
To efficiently generate realistic and precise CXRs from medical reports, we propose a novel disease-knowledge enhanced diffusion-based TTI framework, named Diff-CXR, which takes a deep insight into the characteristics of medical images and reports. Our Diff-CXR achieves SOTA performance on two widely-used report-to-CXR generation benchmarks. Further downstream experiments demonstrate that images generated by Diff-CXR can benefit both traditional thorax disease classification task and multi-modal CXR-report generation task. Our future work will focus on training generative foundation model more efficiently and extending its application to a broader range of tasks.

\section*{Declaration of competing interest}
The authors declare that they have no known competing financial interests or personal relationships that could have appeared to influence the work reported in this paper.

\bibliographystyle{unsrt}
\bibliography{reference}

\begin{thebibliography}{10}

\bibitem{MaPathSRGAN2020}
Jiabo Ma, Jingya Yu, Sibo Liu, Li~Chen, Xu~Li, Jie Feng, Zhixing Chen, Shaoqun Zeng, Xiuli Liu, and Shenghua Cheng.
\newblock Pathsrgan: Multi-supervised super-resolution for cytopathological images using genezative adversarial network.
\newblock {\em IEEE Transactions on Medical Imaging}, 39(9):2920--2930, March 2020.

\bibitem{MingliCTsynthesis2023}
Ming Li, Jiping Wang, Yang Chen, Yufei Tang, Zhongyi Wu, Yujin Qi, Haochuan Jiang, Jian Zheng, and Benjamin M.~W. Tsui.
\newblock Low-dose ct image synthesis for domain adaptation imaging using a generative adversarial network with noise encoding transfer learning.
\newblock {\em IEEE Transactions on Medical Imaging}, 42(9):2616--2630, March 2023.

\bibitem{yeung2024sensorless}
Pak-Hei Yeung, Linde~S Hesse, Moska Aliasi, Monique~C Haak, INTERGROWTH 21st Consortium, Weidi Xie, and Ana~IL Namburete.
\newblock Sensorless volumetric reconstruction of fetal brain freehand ultrasound scans with deep implicit representation.
\newblock {\em Medical Image Analysis}, 94:103147, 2024.

\bibitem{PatrickTransformerSynthesis}
Patrick Esser, Robin Rombach, and Bjorn Ommer.
\newblock Taming transformers for high-resolution image synthesis.
\newblock In {\em 2021 {{IEEE Conference}} on {{Computer Vision}} and {{Pattern Recognition}} ({{CVPR}})}, pages 12873--12883, 2021.

\bibitem{WU2024103284}
Shaoju Wu, Sila Kurugol, and Tsai Andy.
\newblock Improving the radiographic image analysis of the classic metaphyseal lesion via conditional diffusion models.
\newblock {\em Medical Image Analysis}, 97:103284, 2024.

\bibitem{goodfellowGenerativeAdversarialNets2014}
Ian~J. Goodfellow, Jean {Pouget-Abadie}, Mehdi Mirza, Bing Xu, David {Warde-Farley}, Sherjil Ozair, Aaron Courville, and Yoshua Bengio.
\newblock Generative adversarial nets.
\newblock In {\em Proceedings of the 27th {{International Conference}} on {{Neural Information Processing Systems}} - {{Volume}} 2}, {{NIPS}}'14, pages 2672--2680, {Cambridge, MA, USA}, December 2014. {MIT Press}.

\bibitem{arjovskyWassersteinGenerativeAdversarial2017}
Martin Arjovsky, Soumith Chintala, and L{\'e}on Bottou.
\newblock Pathsrgan: Multi-supervised super-resolution for cytopathological images using generative adversarial network.
\newblock In {\em Proceedings of the 34th {{International Conference}} on {{Machine Learning}} - {{Volume}} 70}, {{ICML}}'17, pages 214--223, {Sydney, NSW, Australia}, August 2017. {JMLR.org}.

\bibitem{MehdiCGAN2014}
Mehdi Mirza and Simon Osindero.
\newblock Conditional generative adversarial nets, November 2014.

\bibitem{lee2024vision}
Hyungyung Lee, Da~Young Lee, Wonjae Kim, Jin-Hwa Kim, Tackeun Kim, Jihang Kim, Leonard Sunwoo, and Edward Choi.
\newblock Vision-language generative model for view-specific chest x-ray generation.
\newblock In {\em Conference on Health, Inference, and Learning}, pages 280--296. PMLR, 2024.

\bibitem{llmcxr2024}
Suhyeon Lee, Won~Jun Kim, Jinho Chang, and Jong~Chul Ye.
\newblock Llm-cxr: Instruction-finetuned llm for cxr image understanding and generation.
\newblock In {\em The {{Tenth International Conference}} on {{Learning Representations}}, {{ICLR}} 2024}. {OpenReview.net}, 2024.

\bibitem{RobinStableDiffusion}
Robin Rombach, Andreas Blattmann, Dominik Lorenz, Patrick Esser, and Bj{\"o}rn Ommer.
\newblock High-resolution image synthesis with latent diffusion models.
\newblock In {\em 2022 {{IEEE Conference}} on {{Computer Vision}} and {{Pattern Recognition}} ({{CVPR}})}, pages 12873--12883, 2022.

\bibitem{PierreFewshotSD2022}
Pierre Chambon, Christian Bluethgen, Curtis~P Langlotz, and Akshay Chaudhari.
\newblock Adapting pretrained vision-language foundational models to medical imaging domains, October 2022.

\bibitem{RoentGen2022}
Pierre Chambon, Christian Bluethgen, Jean-Benoit Delbrouck, Rogier Van~der Sluijs, Ma{\l}gorzata Po{\l}acin, Juan Manuel~Zambrano Chaves, Tanishq~Mathew Abraham, Shivanshu Purohit, Curtis~P Langlotz, and Akshay Chaudhari.
\newblock Roentgen: Vision-language foundation model for chest x-ray generation, November 2022.

\bibitem{Antanas2023media}
Antanas Kascenas, Pedro Sanchez, Patrick Schrempf, Chaoyang Wang, William Clackett, Shadia~S Mikhael, Jeremy~P Voisey, Keith Goatman, Alexander Weir, Nicolas Pugeault, Sotirios~A Tsaftaris, and O'Neil~Alison Q.
\newblock The role of noise in denoising models for anomaly detection in medical images.
\newblock {\em Medical Image Analysis}, 90:102963, December 2023.

\bibitem{ZhouAnomaly2021}
Zhou Kang, Li~Jing, Luo Weixin, Li~Zhengxin, Yang Jianlong, Fu~Huazhu, Cheng Jun, Liu Jiang, and Gao Shenghua.
\newblock Proxy-bridged image reconstruction network for anomaly detection in medical images.
\newblock {\em IEEE Transactions on Medical Imaging}, 41(3):582--594, October 2021.

\bibitem{BiomedClip2023}
Sheng Zhang, Yanbo Xu, Naoto Usuyama, Hanwen Xu, Jaspreet Bagga, Robert Tinn, Sam Preston, Rajesh Rao, Mu~Wei, Naveen Valluri, Cliff Wong, Andera Tupini, Yu~Wang, Matt Mazzola, Swadheen Shukla, Lars Liden, Jianfeng Gao, Matthew~P Lungren, Tristan Naumann, Sheng Wang, and Hoifung Poon.
\newblock Biomedclip: A multimodal biomedical foundation model pretrained from fifteen million scientific image-text pairs, March 2023.

\bibitem{huang2024chest}
Peng Huang, Xue Gao, Lihong Huang, Jing Jiao, Xiaokang Li, Yuanyuan Wang, and Yi~Guo.
\newblock Chest-diffusion: A light-weight text-to-image model for report-to-cxr generation.
\newblock In {\em 2024 IEEE International Symposium on Biomedical Imaging (ISBI)}, pages 1--5. IEEE, 2024.

\bibitem{RameshDiffusion2022}
Aditya Ramesh, Prafulla Dhariwal, Alex Nichol, Casey Chu, and Mark Chen.
\newblock Hierarchical text-conditional image generation with clip latents, April 2022.

\bibitem{UVIT2023}
Fan Bao, Shen Nie, Kaiwen Xue, Yue Cao, Chongxuan Li, Hang Su, and Jun Zhu.
\newblock All are worth words: A vit backbone for diffusion models.
\newblock In {\em 2023 {{IEEE Conference}} on {{Computer Vision}} and {{Pattern Recognition}} ({{CVPR}})}, pages 22669--22679, 2023.

\bibitem{zhang2023adding}
Lvmin Zhang, Anyi Rao, and Maneesh Agrawala.
\newblock Adding conditional control to text-to-image diffusion models.
\newblock In {\em Proceedings of the IEEE/CVF International Conference on Computer Vision}, pages 3836--3847, 2023.

\bibitem{radford2021learning}
Alec Radford, Jong~Wook Kim, Chris Hallacy, Aditya Ramesh, Gabriel Goh, Sandhini Agarwal, Girish Sastry, Amanda Askell, Pamela Mishkin, Jack Clark, Gretchen Krueger, and Ilya Sutskever.
\newblock Learning transferable visual models from natural language supervision.
\newblock In {\em International conference on machine learning}, pages 8748--8763. PMLR, 2021.

\bibitem{Pubmedclip2023}
Sedigheh Eslami, Christoph Meinel, and Gerard De~Melo.
\newblock Pubmedclip: How much does clip benefit visual question answering in the medical domain?
\newblock In {\em Findings of the Association for Computational Linguistics: EACL 2023}, page 1181–1193, 2023.

\bibitem{MedClip2022}
Zifeng Wang, Zhenbang Wu, Dinesh Agarwal, and Jimeng Sun.
\newblock Medclip: Contrastive learning from unpaired images and text, October 2022.

\bibitem{GuyuCLIP2021}
Yu~Gu, Robert Tinn, Hao Cheng, Michael Lucas, Naoto Usuyama, Xiaodong Liu, Tristan Naumann, Jianfeng Gao, and Hoifung Poon.
\newblock Domain-specific language model pretraining for biomedical natural language processing.
\newblock {\em ACM Transactions on Computing for Healthcare (HEALTH)}, 3(1):1--23, October 2021.

\bibitem{chexbert}
Akshay Smit, Saahil Jain, Pranav Rajpurkar, Anuj Pareek, Andrew~Y Ng, and Matthew~P Lungren.
\newblock Combining automatic labelers and expert annotations for accurate radiology report labeling using bert.
\newblock In {\em Proceedings of the 2020 Conference on Empirical Methods in Natural Language Processing (EMNLP)}, page 1500–1519, 2020.

\bibitem{mimiccxr}
Alistair~EW Johnson, Tom~J Pollard, Seth~J Berkowitz, Nathaniel~R Greenbaum, Matthew~P Lungren, Chih-ying Deng, Roger~G Mark, and Steven Horng.
\newblock Mimic-cxr, a de-identified publicly available database of chest radiographs with free-text reports.
\newblock {\em Scientific Data}, 6(1):317, November 2019.

\bibitem{IUxray}
Dina Demner-Fushman, Marc~D Kohli, Marc~B Rosenman, Sonya~E Shooshan, Laritza Rodriguez, Sameer Antani, George~R Thoma, and Clement~J McDonald.
\newblock Preparing a collection of radiology examinations for distribution and retrieval.
\newblock {\em Journal of the American Medical Informatics Association}, 23(2):304--310, 2015.

\bibitem{wang2017chestx}
Xiaosong Wang, Yifan Peng, Le~Lu, Zhiyong Lu, Mohammadhadi Bagheri, and Ronald~M Summers.
\newblock Chestx-ray8: Hospital-scale chest x-ray database and benchmarks on weakly-supervised classification and localization of common thorax diseases.
\newblock In {\em Proceedings of the IEEE conference on computer vision and pattern recognition}, pages 2097--2106, 2017.

\bibitem{irvin2019chexpert}
Jeremy Irvin, Pranav Rajpurkar, Michael Ko, Yifan Yu, Silviana Ciurea-Ilcus, Chris Chute, Henrik Marklund, Behzad Haghgoo, Robyn Ball, Katie Shpanskaya, Jayne Seekins, David~A Mong, Safwan~S Halabi, Jesse~K Sandberg, Ricky Jones, David~B Larson, Curtis~P. Langlotz, Bhavik~N Patel, Matthew~P Lungren, and Andrew~Y Ng.
\newblock Chexpert: A large chest radiograph dataset with uncertainty labels and expert comparison.
\newblock In {\em Proceedings of the AAAI conference on artificial intelligence}, volume~33, pages 590--597, 2019.

\bibitem{huang2017densely}
Gao Huang, Zhuang Liu, Laurens Van Der~Maaten, and Kilian~Q Weinberger.
\newblock Densely connected convolutional networks.
\newblock In {\em Proceedings of the IEEE conference on computer vision and pattern recognition}, pages 4700--4708, 2017.

\bibitem{torchxrayvision}
Joseph~Paul Cohen, Joseph~D Viviano, Paul Bertin, Paul Morrison, Parsa Torabian, Matteo Guarrera, Matthew~P Lungren, Akshay Chaudhari, Rupert Brooks, Mohammad Hashir, and Hadrien Bertrand.
\newblock Torchxrayvision: A library of chest x-ray datasets and models, October 2021.

\bibitem{bleu}
Kishore Papineni, Salim Roukos, Todd Ward, and Wei-Jing Zhu.
\newblock Bleu: a method for automatic evaluation of machine translation.
\newblock In {\em Proceedings of the 40th Annual Meeting of the Association for Computational Linguistics}, page 311–318, 2002.

\bibitem{Meteor}
Satanjeev Banerjee and Alon Lavie.
\newblock Meteor: An automatic metric for mt evaluation with improved correlation with human judgments.
\newblock In {\em Proceedings of the ACL Workshop on Intrinsic and Extrinsic Evaluation Measures for Machine Translation and/or Summarization}, page 65–72. Association for Computational Linguistics, June 2005.

\bibitem{Rouge}
Chin-Yew Lin.
\newblock Rouge: A package for automatic evaluation of summaries.
\newblock In {\em Text Summarization Branches Out}, pages 74--81. Association for Computational Linguistics, July 2004.

\bibitem{ChexClusion}
Laleh Seyyed-Kalantari, Guanxiong Liu, Matthew McDermott, Irene~Y Chen, and Marzyeh Ghassemi.
\newblock Chexclusion: Fairness gaps in deep chest x-ray classifiers.
\newblock In {\em BIOCOMPUTING 2021: proceedings of the Pacific symposium}, pages 232--243, 2020.

\bibitem{VITMAE}
Junfei Xiao, Yutong Bai, Alan Yuille, and Zongwei Zhou.
\newblock Delving into masked autoencoders for multi-label thorax disease classification.
\newblock In {\em Proceedings of the IEEE/CVF Winter Conference on Applications of Computer Vision}, pages 3588--3600, 2023.

\bibitem{liu2022convnet}
Zhuang Liu, Hanzi Mao, Chao-Yuan Wu, Christoph Feichtenhofer, Trevor Darrell, and Saining Xie.
\newblock A convnet for the 2020s.
\newblock In {\em Proceedings of the IEEE/CVF conference on computer vision and pattern recognition}, pages 11976--11986, 2022.

\bibitem{tu2022maxvit}
Zhengzhong Tu, Hossein Talebi, Han Zhang, Feng Yang, Peyman Milanfar, Alan Bovik, and Yinxiao Li.
\newblock Maxvit: Multi-axis vision transformer.
\newblock In {\em European conference on computer vision}, pages 459--479. Springer, 2022.

\bibitem{R2GENCMN}
Zhihong Chen, Yaling Shen, Yan Song, and Xiang Wan.
\newblock Cross-modal memory networks for radiology report generation, October 2022.

\bibitem{wang2022cross}
Jun Wang, Abhir Bhalerao, and Yulan He.
\newblock Cross-modal prototype driven network for radiology report generation.
\newblock In {\em European Conference on Computer Vision}, pages 563--579. Springer, 2022.

\end{thebibliography}

\end{document}